\documentclass{article}

\usepackage[final]{corl_2019} 

\usepackage{xspace}
\usepackage{graphicx}
\usepackage{amsmath,amssymb}
\usepackage{color}
\usepackage{multirow}
\usepackage[caption=false]{subfig}
\usepackage{tabulary}
\usepackage{bm}
\usepackage{booktabs}
\usepackage{array}
\usepackage{ragged2e}
\usepackage{enumitem}
\usepackage{capt-of}
\usepackage{xfrac}
\usepackage{floatrow}

\makeatletter
\DeclareRobustCommand\onedot{\futurelet\@let@token\@onedot}
\def\@onedot{\ifx\@let@token.\else.\null\fi\xspace}
\def\eg{\emph{e.g}\onedot} 
\def\ie{\emph{i.e}\onedot} 
 
\def\etc{\emph{etc}\onedot}

\makeatother

\newcolumntype{x}[1]{>{\centering\arraybackslash}p{#1pt}}
\newlength\savewidth

\newcommand{\bbR}{\mathbb{R}}

\newcommand{\bx}{\mathbf{x}}

\newcommand{\tb}{\textbf}
\newcommand{\cm}{\checkmark}

\title{Identifying Unknown Instances for Autonomous Driving}

\author{
\textbf{Kelvin Wong\textsuperscript{1,2}, Shenlong Wang\textsuperscript{1, 2}, Mengye Ren\textsuperscript{1, 2}, Ming Liang\textsuperscript{1}, Raquel Urtasun\textsuperscript{1, 2}} \\
Uber Advanced Technologies Group\textsuperscript{1}, University of Toronto\textsuperscript{2} \\
\texttt {\{kelvin.wong, slwang, mren3, ming.liang, urtasun\}@uber.com}
}

\begin{document}
\maketitle


\begin{abstract}
In the past few years, we have seen great progress in perception algorithms,
particular through the use of deep learning.
However, most existing approaches focus on a few categories of interest,
which represent only a small fraction of the potential categories that
robots need to handle in the real-world.
Thus, identifying objects from unknown classes remains a challenging yet
crucial task.
In this paper, we develop a novel open-set instance segmentation algorithm
for point clouds which can segment objects from both known and unknown classes
in a holistic way.
Our method uses a deep convolutional neural network to project points into
a category-agnostic embedding space in which they can be clustered into instances
irrespective of their semantics.
Experiments on two large-scale self-driving datasets validate the effectiveness
of our proposed method.
\end{abstract}

\section{Introduction}
One relaxing summer weekend, I drove my family on an excursion to the zoo. All of a sudden, I
saw a tiny black creature in front of my car. It was too far away for me to tell what it was, but as an
experienced driver, I rolled out a series of moves without hesitation: I performed a shoulder check,
I signaled, and then I switched lanes.
In the end, I still had not figured out what it was until my daughter
told me it was a raccoon crossing the street.
The ability to recognize an object without knowing its semantics
seems innate to us humans. However, this is in fact one of the holy grails
that we strive to develop in our robotic perception systems.

A common paradigm in robotics perception is to train and deploy a machine-learned model under the
\textit{closed-set} condition; \ie, the robot is only trained to identify instances from known
classes. In this paper, we argue that this is not enough for a practical perception system, since in
real-world applications, robots often have to operate in an open environment interacting with
surrounding objects that were not seen during training.
Thus, an ideal perception system should be capable of recognizing and localizing
objects from both known and unknown classes.
This is referred to as the \textit{open-set} setting.

We are not the first to realize the importance of identifying and interacting with unknown
instances. In the pioneering work by Saxena et al.~\cite{novelgrasp}, the authors proposed to grasp
a novel object by identifying good positions to grasp; this could be trained on known
instances and then generalized to unknowns. Also, cognitive scientists
have studied
the underlying mechanism
by which the human vision system detects novel objects. In the 1980s, experiments on novel objects
were conducted on rats to reveal how long/short term memory influences
recognition~\cite{ennaceur1988new, antunes2012novel}. In the computer vision community, researchers approached the
open-set recognition problem by first defining the open space as the space sufficiently far from any
known positive training samples, measured by a multi-class classification function, where unknowns
would carry all zero values in the classifier outputs \cite{openworld, bendale2016towards}. However, these approaches
are either restricted to a classification task or specific to downstream robotics tasks. We
generalize this idea to open-set instance segmentation with the additional capability to
group observations into the same instance.

\begin{figure}[h]
\includegraphics[width=\linewidth]{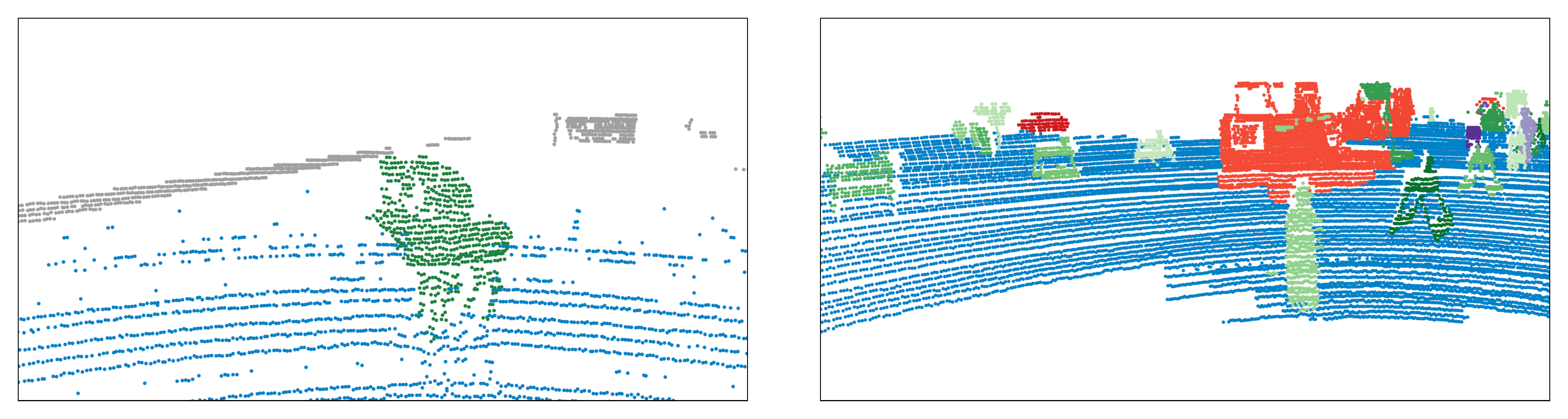}
\caption{
    Exemplar output of open-set instance segmentation.
    The left figure depicts a horse on the road while the right figure depicts
    several construction elements.
    These objects are uncommon sights in day-to-day driving.
    However, an ideal robotics perception system should still be able to
    recognize and localize such objects, and determine whether they belong to
    one of the known classes.
}
\label{fig:teaser}\vspace{-0.1in}
\end{figure}

Recognizing and segmenting an object without seeing its category during training is fundamentally
challenging for modern deep networks. It makes the networks unable to exploit shape, appearance, and
other information about the category during training. However the aforementioned capability
critically influences deep models' success. Back to the mid-20th century, vision scientists identified a
mechanism in the human vision system which groups visual elements that belong together into an
object. This mechanism is called \textit{perceptual grouping} \cite{palmer2002organizing} and it
contributes to our ability to recognize novel objects.
Motivated by the success of the human vision system, our goal is to empower robots with a similar
capability; \ie, we would like them to learn to perceptually group visual elements into a ``thing'',
and then classify whether it belongs to one of the known classes.

Towards this goal of jointly recognizing both known and unknown instances,
we propose a novel perception algorithm for LiDAR point clouds: the Open-Set Instance Segmentation (OSIS) network.
The high-level idea is to use a deep convolutional network to identify uncertain points and
group them into novel unknown instances. Specifically, we propose a category-agnostic instance embedding
network to project points from the same instance to be close together in the embedding
space. As a result, the network learns to group observations
into a thing without knowing the thing's category. Our open-set inference procedure is
straightforward: We first compute prototypical features from each known-class instance and
then associate points with them according to their embedding feature distance.
Finally, we cluster the rest of the points to form new unknown-class instances.

We validate our model's performance on two large-scale self-driving datasets with unknown
objects.
Our experiments show that OSIS outperforms other competing methods in terms of
identifying instances from both known and unknown classes.
\section{Related Work}

The problem of segmentation originates from the concept of perceptual grouping \cite{palmer1999vision},
which argues that human perception organizes perceptual signals into objects and meaningful clusters
instead of raw pixels.
Early segmentation approaches \cite{ncut,graphbased} mainly deal
with low-level regions and often do not capture the notion of objects. Recently, with the growing
availability of high-quality segmentation labels, several benchmarks~\cite{voc,mscoco,cityscapes} have become very
popular for both semantic and instance segmentation tasks.
Panoptic segmentation~\cite{panoptic} was proposed to combine the two problems together by
jointly reasoning about instances and background.

\looseness=-1
Standard instance and panoptic segmentation approaches, however, fail to capture unknown instances
that have never appeared in the training set. Towards the goal of explaining all pixels in the
scene, open-set or out-of-distribution detection has been studied in the classification settings
\cite{longtail,confcalib}. Typically a threshold is learned such that predictions below the
threshold are classified as unknown. \cite{confcalib} proposed to use generative models to help
calibrate the confidence level. Open-set recognition is also closely related to zero-shot learning
\cite{devise,zslcrossmodal,latentzsl}; however, the latter puts more emphasis on 
bootstrapping novel concepts from cross modality inputs (\eg natural languages).

\looseness=-1
Recently, several approaches have been proposed to address the open-set instance segmentation scenario.
\cite{bayesseg} proposes a Bayesian framework that combines an instance segmentation network
\cite{maskrcnn} for known classes with an off-the-shelf contour detection algorithm \cite{ucm} for
unknown classes. This approach can potentially be limited by the capacity of the offline contour
detection algorithm. \cite{segeverything} extends the standard instance segmentation task with
thousands of extra visual concepts in the form of weak labels~\cite{visualgenome}, covering a wide
range of rare objects. This is, however, still closed-set recognition with weak labels. In
\cite{generic4d}, a category-agnostic object proposal network is trained and applied on video
sequences. Due to its ``proposal + classification'' nature, the model may learn to suppress unknown
objects that are present but not labeled in the training examples. In the 3D point clouds domain,
\cite{trevor2013efficient} proposed to leverage connected components, which could be less robust to
cluttered scenes.

Next we review existing literature on instance segmentation. One mainstream approach for instance
segmentation is based on object detection boxes~\cite{maskrcnn,recattend,upsnet}, where object
segmentations are produced within detection boxes.
These approaches are referred to as ``two-stage'' joint detection and segmentation models.
\cite{tensormask,deepmask,sharpmask} output object instance proposals directly
from each pixel. For 3D point cloud, \cite{deepsliding,fpointnet} also use similar two-stage
architectures to perform point cloud detection and segmentation. As segmentation happens after
detection, unknown objects are often left unrecognized. Unless the object detector is trained to
recognize unknown classes, these approaches are likely unsuitable for our open-set instance
segmentation problem.

Another line of work for instance segmentation is based on bottom-up grouping of pixels. \cite{dwt}
predicts energy of each instances and obtains instance segmentations using flood fill.
\cite{proposalfree,jointbandwidth} cluster pixels by their predicted centroid locations. \cite{sgn}
predicts breaking points on vertical and horizontal directions, and segments objects using line
scanning. In 3D point cloud instance segmentation, several bottom-up approaches have also been
proposed. \cite{sgpn} predicts point affinity to make segmentation proposals for each point. When
the number of points is large, as in the case of LiDAR point clouds, this approach can end up with
too many proposals to process.

Our method is most similar to a line of work in bottom-up segmentation that learns instance-aware
embeddings~\cite{pushpull,masc,jsis,bevis,metricseg,recurrentemb}. Our method and these approaches
all use clustering (\eg mean-shift~\cite{meanshift}, DBSCAN~\cite{dbscan}, \etc) to aggregate the
points into instances based on their embedding similarities. 
Despite having a similar instance-aware embedding component, our proposed
method is distinguished by two major differences.
First, we leverage an object detection head to propose anchors against which
points can be clustered, thus resulting in a more efficient and effective
algorithm with top-down guidance.
Second, we propose to directly predict prototypical features for each anchor
to account for the spatial sparsity and non-uniform density LiDAR point clouds.


\begin{figure}
\centering
\vspace{-0.2in}
\includegraphics[width=0.95\textwidth]{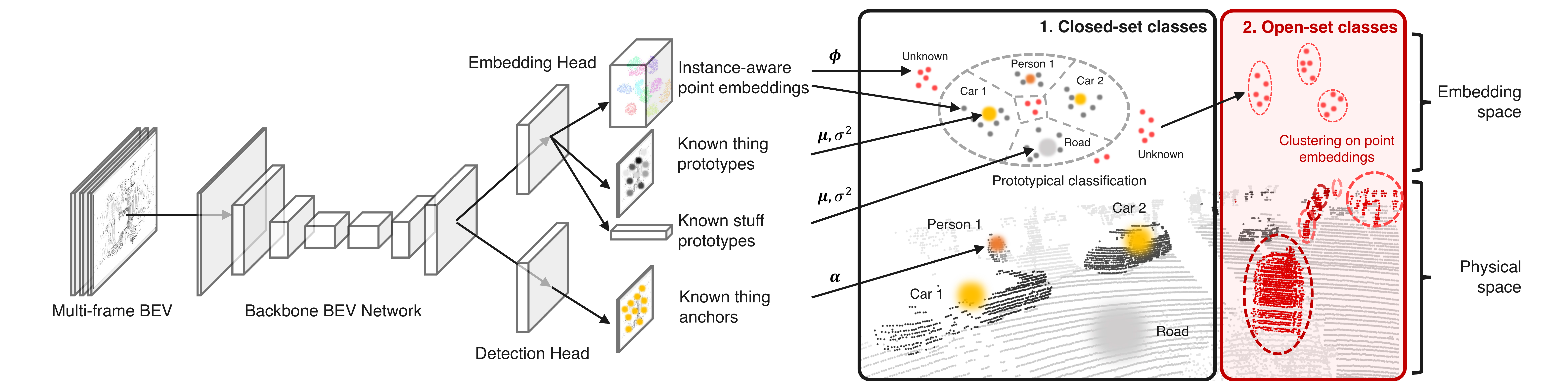}
\caption{
    Our OSIS model contains two branches:
    a) a detection head to detect anchors representing instances of known thing classes; and
    b) an embedding head to extract instance-aware embeddings for each point
    as well as prototypes for each thing anchor and each stuff class.
    In the first stage of inference, the prototypes collectively filter out points from the known classes.
    In the second stage, we cluster the remaining unknown points into instances
    using their embeddings and 3D locations.
}
\label{fig:inference}
\end{figure}

\section{Identifying the Unknowns}
In this paper, we propose the Open-Set Instance Segmentation (OSIS) network
for identifying known and unknown objects from point clouds.
In the following, we first formally define the problem of open-set
instance segmentation in Sec.~\ref{sec:formulation}.
Then, we discuss our full inference framework in Sec.~\ref{sec:approach}.
Finally, we provide details on how to train our model in Sec.~\ref{sec:learning}.

\subsection{Problem Formulation}
\label{sec:formulation}
Let $ \mathcal{X} = \{\bx_i\}_{i = 1}^{N} $ be an input set of $ N $ points,
where each $ \bx_i \in \bbR^D $ is the input feature for point $ i $.
Given a set of instance ids $ \mathbb{I} $ and a set of open-set semantic labels
$ \mathbb{O} $, we want a function $ f $ mapping each input feature
$ \bx_i \in \mathcal{X} $ to a tuple $ (y_i, z_i) \in \mathbb{I} \times \mathbb{O} $.
Note that $ \mathbb{O} $ may be partitioned into two disjoint subsets
$ \mathbb{C} $ and $ \{\bot\} $, where $ \mathbb{C} $ is the set of known
classes and $ \bot $ is the semantic label for the unknown class.
The known classes $ \mathbb{C} $ can be further divided into
$ \mathbb{C}_{\mathrm{thing}} $ and $ \mathbb{C}_{\mathrm{stuff}} $, which
correspond to the known thing classes (\eg, vehicle and pedestrian) and the
known stuff classes (\eg, road) respectively.
As in~\cite{panoptic}, we require that every point with the same instance id
have the same semantic label.
Furthermore, we ignore the instance ids of stuff points.

Our problem formulation differs from standard panoptic segmentation
\cite{panoptic} with regards to how the unknown (void) class is handled.
In the standard setting, we do not require instance labels for points with a
void semantic label.
By contrast, in our setting, we want to identify individual instances
for the unknown class as well.
Fig.~\ref{fig:teaser} shows an example output for this task.

\subsection{Open-Set Instance Segmentation}\label{sec:approach}
In this subsection, we describe our proposed approach for open-set instance
segmentation.
Our approach is based on learning a category-agnostic embedding space
in which points can be clustered into instances irrespective of their semantics.
To this end, we design a convolutional neural network that consists of three
components:
1) a shared backbone feature extractor;
2) a detection head to detect anchors representing instances of known things; and
3) an embedding head to predict instance-aware features for each point
as well as prototypes for each thing anchor and stuff class.

Our inference procedure consists of two stages.
First, we perform closed-set perception by associating points to prototypes of 
known things and stuff using the learned embedding space.
Next, we perform open-set perception by classifying points with uncertain
associations as unknown, and then clustering them into instances using their
instance-aware embeddings and 3D coordinates as features.
We refer the reader to Fig.~\ref{fig:inference} for an illustration of our full
inference pipeline.

\vspace{-0.1in}
\paragraph{Input representation:}
Our model takes as input a bird's eye view (BEV) rasterized image of a LiDAR
point cloud $ \mathcal{X} = \{(\mathrm{x}_i, \mathrm{y}_i, \mathrm{z}_i)\}_{i = 1}^{N} $
centered on the ego-car.
Specifically, we voxelize $ \mathcal{X} $ into a 3D occupancy grid using
reversed trilinear interpolation~\cite{segcloud} and treat its vertical axis as
multi-dimensional features.
This yields a compact yet effective representation of $ \mathcal{X} $ on which
we can use 2D convolutions~\cite{chris}.
Our model can also exploit temporal information by taking multiple BEV LiDAR
frames stacked along the feature channel as input.
To alleviate misalignment across frames due to the ego-car's movements,
we use localization to compensate the ego-motion.

\vspace{-0.1in}
\paragraph{Backbone network:} 
We use a custom 2D convolutional feature pyramid network to extract multi-scale
features from the input BEV LiDAR frame.
In this network, we stack several residual blocks to compute a
feature hierarchy consisting of three scales of the input resolution:
$ 1/4 $, $ 1/8 $, and $ 1/16 $.
These multi-scale features are then upsampled to the $ 1/4 $ scale and fused
via residual connections to output a $ C \times H \times W $ feature map,
where $ C $ is the number of feature channels, and $ H $ and $ W $ is the height
and width of the feature map respectively.
This feature map is subsequently used as input to the detection and embedding heads.

\vspace{-0.1in}
\paragraph{Detection head:}
Our detection head consists of four $ 3 \times 3 $ convolution layers,
followed by a $ 1 \times 1 $ convolution layer.
For each BEV pixel and for each class in $ \mathbb{C}_{\mathrm{thing}} $,
it predicts $ (\alpha, dx, dy, w, l, \sin (2\theta), \cos(2 \theta)) $,
where $ \alpha $ is the anchor confidence score, $ (dx, dy) $ is the position offsets
to its object center, and the rest parameterize the geometry of its bounding
box~\cite{hdnet}.
During inference, we remove anchors with scores less than $ \tau $ to obtain
the set of anchors $ \mathcal{A}_\tau $.
Note that the bounding box parameters are predicted only to exploit additional
supervision signals.

\vspace{-0.1in}
\paragraph{Embedding head:}
The embedding head forms the core of our open-set instance segmentation model:
it learns a category-agnostic embedding space in which points can be
clustered into instances irrespective of their semantics.
Specifically, the embedding head is a four-layer CNN with $ 3 \times 3 $ filters
followed by three distinct branches:

\vspace{-0.1in}
\begin{enumerate}[leftmargin=*]

\looseness=-1
\item The point branch computes features
$ \Phi_{\mathrm{point}} \in \mathbb{R}^{(F \times Z) \times H \times W} $ via
a $ 1 \times 1 $ convolution, where $ F $ is the dimension of the embedding
space, and $ Z $ is the number of bins along the gravitational $ z $-axis.
For each point $ i $ in $ \mathcal{X} $, we extract an embedding $ \bm\phi_i $
from $ \Phi_{\mathrm{point}} $ via trilinear interpolation.

\item The thing branch computes features
$ \Phi_{\mathrm{thing}} \in \mathbb{R}^{(F + 1) \times H \times W} $
via a $ 1 \times 1 $ convolution.
For each anchor $ k $ in $ \mathcal{A}_\tau $, we extract its
prototype $ (\bm{\mu}_k, \sigma^2_k) \in \mathbb{R}^F \times \mathbb{R} $
by bilinearly interpolating $ \Phi_{\mathrm{thing}} $ around the anchor's object center.
This yields a set of thing prototypes $ \mathcal{P}_{\mathrm{thing}} $.

\item The stuff branch performs global average pooling to obtain features
$ \Phi_{\mathrm{stuff}} \in \mathbb{R}^{C \times 1 \times 1} $.
For each stuff class $ c \in \mathbb{C}_{\mathrm{stuff}} $, we apply a linear layer
on $ \Phi_{\mathrm{stuff}} $ to predict its prototype
$ (\bm{\mu}_c, \sigma^2_c) \in \mathbb{R}^F \times \mathbb{R} $.
This yields a set of stuff prototypes $ \mathcal{P}_{\mathrm{stuff}} $.
\end{enumerate}

\vspace{-0.1in}
\paragraph{Closed-set perception:}
Our closed-set perception algorithm draws inspiration from prototypical networks
for few-shot learning~\cite{protonet}.
First, we apply non-maximum suppression to $ \mathcal{P}_{\mathrm{thing}} $
to obtain a unique set of thing prototypes $ \mathcal{P}_{\mathrm{thing}}' $.
Let us denote $ \mathcal{P}_{\mathrm{all}} = \mathcal{P}_{\mathrm{thing}}' \cup \mathcal{P}_{\mathrm{stuff}} $
as the final set of all thing and stuff prototypes.
Then, given a point $ i $ in $ \mathcal{X} $, we compute its point-to-prototype
association score with respect to every prototype $ k $ in
$ \mathcal{P}_{\mathrm{all}} $ as follows:
\begin{align}
    \bm{\hat{y}}_{i, k} = {-\frac{\|\bm\phi_i - \bm\mu_k\|^2}{2 \sigma_k^2} - \frac{F}{2} \log \sigma^2_k}
\end{align}

Additionally, we have a learnable global constant $ U $ corresponding to its
score $ \bm{\hat{y}}_{i, |\mathcal{P}_{\mathrm{all}}| + 1} $ of not associating with any prototype in
$ \mathcal{P}_{\mathrm{all}} $.
Thus, its instance label can be computed by taking the argmax over its
association scores $ \bm{\hat{y}}_i $.
Furthermore, its semantic label is simply the class of its instance, or
unknown if it is not associated with any prototypes in
$ \mathcal{P}_{\mathrm{all}} $.
Note that, in practice, we compute each point's scores only with the
prototypes of its $ k $-nearest thing anchors and all
$ |\mathbb{C}_{\mathrm{stuff}}| $ stuff classes;
this helps to accelerate inference speed.

\vspace{-0.1in}
\paragraph{Identifying unknown instances:}
We assign instance labels to unknown points via DBSCAN~\cite{dbscan} clustering.
Specifically, for two points $ \bm{x_i}, \bm{x_j} \in \mathcal{X} $, their
pairwise distance used in DBSCAN is a convex combination of their point
embedding squared distance and their 3D location squared distance; \ie, 
\begin{align}
    d_{ij}^2 = \beta \|\bm{x_i} - \bm{x_j}\|^2 + (1 - \beta) \|\bm{\phi}_i - \bm{\phi}_j\|^2
\end{align}

Combining the instance labels obtained from this stage with the results from
closed-set perception, we obtain our final open-set instance segmentation
predictions.

\begin{figure}[t]
    \includegraphics[width=\textwidth]{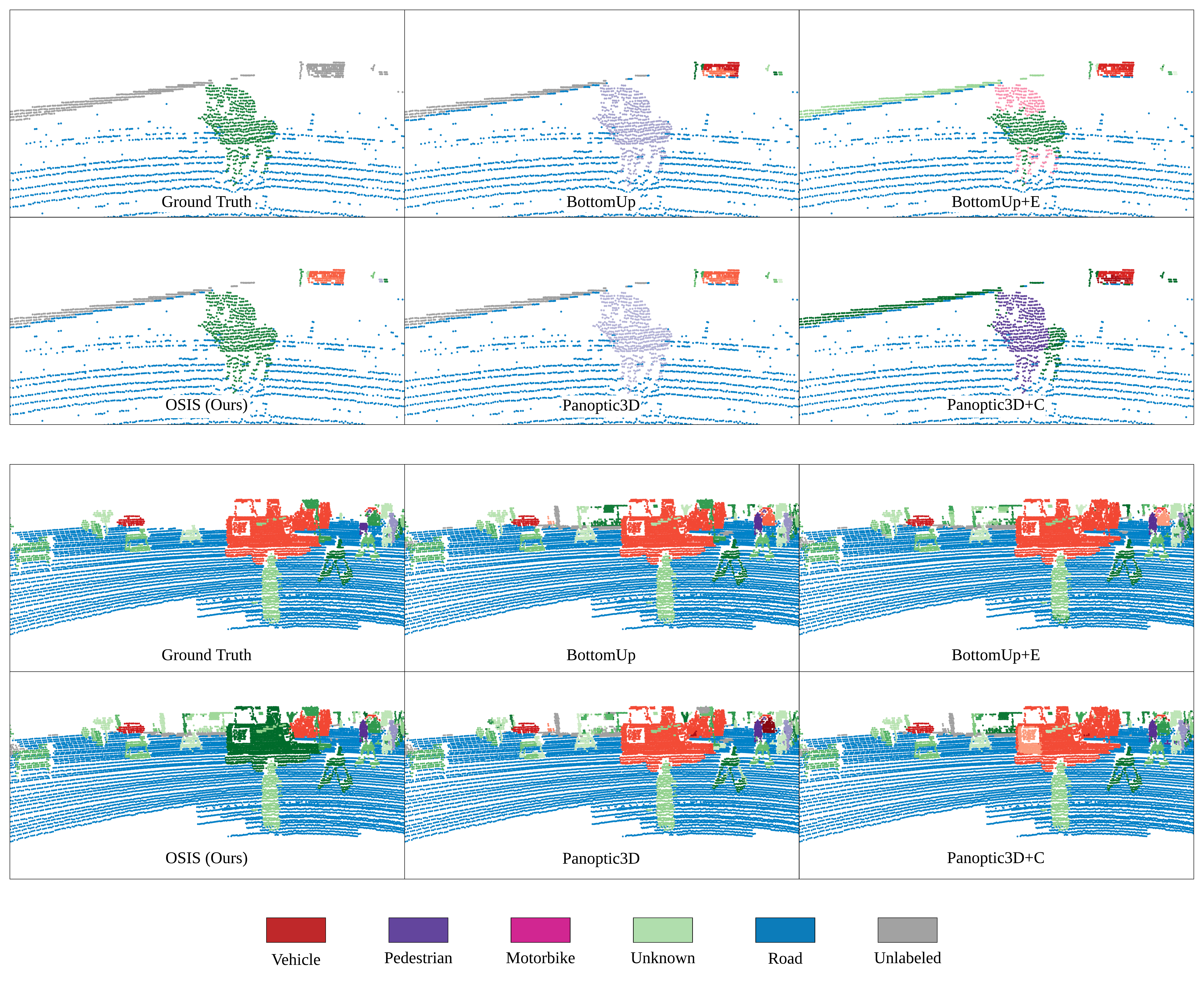}
    \caption{
        Qualitative results of open-set instance segmentation on TOR4D and Rare4D.
    }
    \label{fig:results}
\end{figure}

\subsection{Learning}
\label{sec:learning}

Our model is optimized with respect to a combination of detection and embedding
losses:
\begin{align}
    \mathcal{L} = \lambda_{\mathrm{det}} \ell_{\mathrm{det}} +
                  \lambda_{\mathrm{emb}} \ell_{\mathrm{emb}}
\end{align}
where $ \ell_{\mathrm{det}} $ is the detection loss, $ \ell_{\mathrm{emb}} $ is
the embedding loss, and $ \lambda $'s are their associated loss weights.
In our experiments, we set $ \lambda $'s to 1.
Since $ \mathcal{L} $ is fully differentiable with respect to the network parameters,
we train our model using the standard back-propagation algorithm.

\vspace{-0.1in}
\paragraph{Detection loss:}
We use a standard multi-task loss function to train the detection head.
In particular, for object classification, we use binary cross-entropy with
online negative hard mining, where positive and negative BEV pixels are
determined by their distances to an object center~\cite{hdnet}.
For bounding box regression, we use a combination of IoU loss for box locations
and sizes and SmoothL1 loss for box orientations on predictions at positive pixels.
It is worth noting that box sizes and orientations are not used during inference,
and we predict them only for a stronger supervision signal.

\vspace{-0.1in}
\paragraph{Embedding loss:}
We use a standard cross-entropy loss function to encourage points to be assigned
to the correct prototype.
In particular, during training we first gather a set of prototypes
$ \mathcal{P}_{\mathrm{gt}} $, which is the union of $ \mathcal{P}_{\mathrm{stuff}} $
and the set of thing prototypes obtained by bilinearly interpolating
$ \Phi_{\mathrm{thing}} $ around ground truth object centers.
Next, we compute point-to-prototype association scores
$ \{\bm{\hat{y}}_i\}_{i = 1}^{N} $ with respect to $ \mathcal{P}_{\mathrm{gt}} $,
and normalize each $ \bm{\hat{y}}_i $ using the softmax function.
Finally, we calculate the cross-entropy loss as follows:
\begin{align}
    \ell_{\mathrm{proto}} = -\frac{1}{N}\sum_{i = 1}^{N} 
        \sum_{k = 1}^{|\mathcal{P}_{\mathrm{gt}}| + 1} \bm{y}_{i, k} \log \bm{\hat{y}}_{i, k}
\end{align}
where each $ \bm{y}_i $ is a one-hot vector indicating ground truth associations.
We also apply a discriminative loss function~\cite{pushpull} on the point
embeddings $ \{\bm{\phi}_i\}_{i = 1}^{N} $, which we found improves performance.

\newfloatcommand{capbtabbox}{table}[][\FBwidth]

\begin{table}[!tbp]
    \captionsetup{size=footnotesize}
    \centering
    \resizebox{\textwidth}{!}{%
    \begin{tabular}{cccc|ccc|cccccc}
    \toprule
                                                 & \multicolumn{3}{c|}{Rare4D}        & \multicolumn{9}{c}{TOR4D}\\
                                                   \cmidrule{2-4}                       \cmidrule{5-7}                       \cmidrule{8-10}                     \cmidrule{11-13}
                                                 & \multicolumn{3}{c|}{Unknown}       & \multicolumn{3}{c|}{Unknown      } & \multicolumn{3}{c}{Known Thing}   & \multicolumn{3}{c}{Known Stuff}\\
                                                   \cmidrule{2-4}                       \cmidrule{5-7}                       \cmidrule{8-10}                     \cmidrule{11-13}
                                                 & UQ        & RQ        & SQ         & UQ        & RQ        & SQ         & PQ        & RQ        & SQ        & PQ        & RQ    & SQ         \\
    \midrule
    MT-PNet~\cite{jsis}                          & 10.9      & 11.9      & 91.9       & 27.7      & 31.6      & 87.8       & 27.6      & 29.4      & 91.9      & 94.4      & 100.0 & 94.4       \\
    \midrule
    BottomUp~\cite{chris}                        & 48.7      & 53.7      & 90.7       & 49.2      & 54.8      & 89.8       & 56.0      & 60.1      & 92.8      & \tb{98.2} & 100.0 & \tb{98.2}  \\
    BottomUp+E~\cite{chris}                      & 57.9      & 62.8      & 91.0       & 62.7      & 69.1      & 90.7       & 64.1      & 67.2      & 94.9      & \tb{98.2} & 100.0 & \tb{98.2}  \\
    Panoptic3D~\cite{panoptic}                   & 41.7      & 49.2      & 84.8       & 43.8      & 51.9      & 84.4       & 74.5      & 77.2      & 96.3      & \tb{98.2} & 100.0 & \tb{98.2}  \\
    Panoptic3D+C~\cite{panoptic}                 & 50.8      & 56.2      & 90.3       & 51.5      & 57.3      & 89.9       & 78.8      & 81.4      & \tb{96.6} & 97.9      & 100.0 & 97.9       \\
    \midrule
    OSIS (Ours)                                  & \tb{62.5} & \tb{66.5} & \tb{94.0}  & \tb{66.0} & \tb{71.3} & \tb{92.5}  & \tb{81.5} & \tb{84.3} & \tb{96.6} & 97.7      & 100.0 & 97.7       \\
    \bottomrule
    \end{tabular}%
    }
    \caption{Quantitative results of open-set instance segmentation on the TOR4D and Rare4D test sets.}
    \label{table:open-set-results}
\end{table}

\section{Experiments}
In this section, we showcase the effectiveness of our proposed model OSIS
on two large-scale self-driving datasets.
We first describe our experimental setup and then discuss the results we obtained.

\subsection{Experimental Setup}
\label{sec:experimental-setup}
\paragraph{Datasets:}

\begin{itemize}[leftmargin=*]
\item \textbf{TOR4D}~\cite{hdnet} is a large-scale self-driving dataset collected
from cities across North America.
This dataset consists of 6500 distinct driving scenarios, each containing 250
sweeps of LiDAR point clouds.
We partition TOR4D into a training set of 5000 scenarios,
a validation set of 500, and a test set of 1000. 
Furthermore, we subsample every five frames across all three splits.

Each frame in TOR4D is annotated with per-point semantic and instance labels
according to four classes: vehicle, pedestrian, motorbike, and road.
Points not belonging to one of those classes are unlabled and regarded as unknown.
To evaluate OSIS in the open-set setting, we annotate 5,702 and 10,127 unique
unknown objects with instance labels in the validation and test sets
respectively.

\item \textbf{Rare4D} is a dataset of curated self-driving scenarios containing
289 unique rare objects such as forklifts, tractors, and even horses
(see Fig.~\ref{fig:teaser}).
In our experiments, Rare4D is not used for training but for evaluation
of unknown object identification only.

\end{itemize}

\paragraph{Evaluation metrics:}
For known classes, we report the panoptic quality (PQ), recognition quality
(RQ), and segmentation quality (SQ) metrics proposed in~\cite{panoptic}.
Since the labels in our dataset consider only things that are removeable
to be separate objects (\eg, flags attached to a building will not be labeled),
we decide not to measure precision; instead, we modify PQ into the 
\emph{unknown quality} (UQ), a recall-based metric that measures performance
on annotated instances only:
\begin{align}
    \mathrm{UQ} =
    \underbrace{\frac{\sum_{(p, g) \in \mathrm{TP}} \mathrm{IoU}(p, g)}{\vert\mathrm{TP}\vert}}_{\text{segmentation quality}~(\mathrm{SQ})}
    \times \underbrace{\frac{\vert\mathrm{TP}\vert}{\vert\mathrm{TP}\vert + \vert\mathrm{FN}\vert}}_{\text{recall quality}~(\mathrm{RQ})}
\end{align}
where $ \mathrm{TP} $ is the set of true positives and $ \mathrm{FN} $ is the
set of false negatives.
As in~\cite{panoptic}, a predicted unknown instance $ p $ matches with the
ground truth unknown instance $ g $ if and only if their intersection over
union exceeds 0.5.

\paragraph{Baselines:}
Due to a lack of prior work in open-set instance segmentation for point clouds,
we adapt several deep learning based instance segmentation algorithms to the
open-set setting to serve as baselines.
Note that all baselines except for MT-PNet use the same backbone network
and input representations.

\begin{itemize}[leftmargin=*]

\looseness=-1
\item \textbf{MT-PNet}~\cite{jsis} is a state-of-the-art joint 3D semantic and 
instance segmentation algorithm\footnote{The CRF post-processing stage is not included.}.
We adapt MT-PNet to the open-set setting as follows: 1) we augment its semantic
header to predict an additional unknown class; and 2) we use DBSCAN~\cite{dbscan}
to cluster unknown points into instances based on their embedding distances.

\item \textbf{BottomUp} first runs a state-of-the-art point cloud semantic
segmentation algorithm~\cite{chris} with an additional unknown class, and then
uses DBSCAN~\cite{dbscan} to cluster points of the same class into instances.
We evaluate two versions of this baseline:
1) \textbf{BottomUp} clusters points using their 3D locations; and
2) \textbf{BottomUp+E} clusters points using embeddings learned via a
discriminative loss function~\cite{pushpull}.

\item \textbf{Panoptic3D} is similar to the pioneering panoptic segmentation
algorithm proposed for 2D images~\cite{panoptic}.
We first perform 3D detection and segmentation, and then apply heuristics to
merge the outputs into a panoptic segmentation of the scene.
Unlike~\cite{panoptic}, we train a single network with both a
3D detection and a semantic segmentation header.
Similar to BottomUp, Panoptic3D also predicts an additional unknown class.
We compare two versions of this baseline:
1) \textbf{Panoptic3D} performs class-agnostic detections; and
2) \textbf{Panoptic3D+C} performs class-aware detections and uses DBSCAN to cluster
unknown points into instances based on their 3D locations.

\end{itemize}

\paragraph{Implementation details:}
In our BottomUp, Panoptic3D, and OSIS experiments, we use a
$ 160 \times 160 \times 5 $ meters region of interest centered on the ego-vehicle.
Points within this area are rasterized into a BEV image using reversed trilinear
interpolation~\cite{segcloud} at a discretization resolution of $ 0.15625 $.
We use five frames of LiDAR as input and align them using ego-motion.
This yields an input tensor of size
$ C \times H \times W = 160 \times 1024 \times 1024 $.
We use the Adam optimizer~\cite{adam} with a batch size of 32 and an
initial learning rate of $ 4\mathrm{e}{-3} $, which we decay by 0.1 after every
five epochs for a total of ten epochs.
Note that experiments for MT-PNet follow a similar setup, with the exception
that we feed raw LiDAR point clouds as input to the model.

\subsection{Results}
As shown in Tab.~\ref{table:open-set-results}, OSIS outperforms
the baselines on known and unknown things on all metrics across both datasets.
Our method is also comparable to state-of-the-art semantic segmentation
models for known stuff classes.
Interestingly, BottomUp+E is the best baseline for unknown things
while Panoptic3D+C is the best baseline for known things.
As our results suggest, OSIS achieves the best of both worlds by marrying a
bottom-up approach with top-down guidance.

Qualitative results in Fig.~\ref{fig:results} further higlight our method's
ability to correctly segment instances from both known and unknown classes.
In particular, OSIS is the sole method that correctly segmented
the horse and identified it as an unknown object; by contrast, the baseline
methods suffer from misclassification errors and noise in instance segmentation.
We also illustrate a failure case in the second row of Fig.~\ref{fig:results}.
In this figure, OSIS misclassified a construction vehicle as unknown.
Despite this mistake, our method still successfully segmented the vehicle
as an instance.

\subsection{Ablation Studies}

\begin{table}[!tbp]
    \captionsetup{size=footnotesize}
    \centering
    \begin{tabular}{ccc|ccc|cccccc}
    \toprule
    DL  & BR  & $\sigma^2$ & \multicolumn{3}{c|}{Unknown}        & \multicolumn{3}{c}{Known Things}  & \multicolumn{3}{c}{Known Stuff}   \\
    \cmidrule{1-3}           \cmidrule{4-6}                        \cmidrule{7-9}                      \cmidrule{10-12}
        &     &            & UQ        & RQ        & SQ          & PQ        & RQ        & SQ        & PQ        & RQ        & SQ        \\
    \midrule
        &     &            & 59.8      & 68.4      & 87.5        & 78.2      & 81.9      & 95.0      & \tb{97.7} & 100.0 & \tb{97.7}     \\
    \cm &     &            & 68.1      & 73.2      & 93.0        & 80.6      & 83.9      & 95.8      & \tb{97.7} & 100.0 & \tb{97.7}     \\
    \cm & \cm &            & 68.5      & \tb{73.8} & 92.8        & 81.9      & 85.0      & 96.2      & \tb{97.7} & 100.0 & \tb{97.7}     \\
    \cm & \cm & \cm        & \tb{68.6} & 73.7      & \tb{93.1}   & \tb{82.6} & \tb{85.5} & \tb{96.5} & \tb{97.7} & 100.0 & \tb{97.7}     \\
    \bottomrule
    \end{tabular}%
    \caption{Ablation study of model components on TOR4D validation set.}
    \label{tab:model-choices}
\end{table}

\paragraph{Model design choices:}
We first conduct an ablation study on three components of our model:
1) whether we optimize the discriminative loss (\textbf{DL});
2) whether we perform bounding box regression (\textbf{BR}); and
3) whether we predict per-prototype scalar variances ($ \bm{\sigma^2} $).
Tab.~\ref{tab:model-choices} shows our results on the TOR4D validation set.
From this table, we can see that all three components contribute
towards the overall performance of our model.

\begin{figure}
\captionsetup{size=footnotesize}
\begin{floatrow}
\capbtabbox{%
    \begin{tabular}{cccc}
    \toprule
                             & \multicolumn{3}{c}{Unknown w/ Oracle}         \\
                               \cmidrule{2-4}
                             & UQ        & RQ        & SQ                    \\
    \midrule
    Points                   & 58.1      & 63.9      & 91.0                  \\
    Center                   & 73.7      & 78.6      & \tb{93.8}             \\
    Semantic                 & 16.1      & 18.9      & 85.3                  \\
    \midrule
    Embedding                & 70.3      & 75.8      & 92.8                  \\
    Embedding + Points       & \tb{74.3} & \tb{79.6} & 93.3                  \\
    \bottomrule
    \end{tabular}%
}{%
    \caption{
        Unknown instance segmentation performance using different
        features for clustering. "w/ oracle" indicates that we use
        ground truth to remove known points prior to DBSCAN.
    }
    \label{tab:ablation-features}
}
\ffigbox{%
    \includegraphics[width=0.5\textwidth]{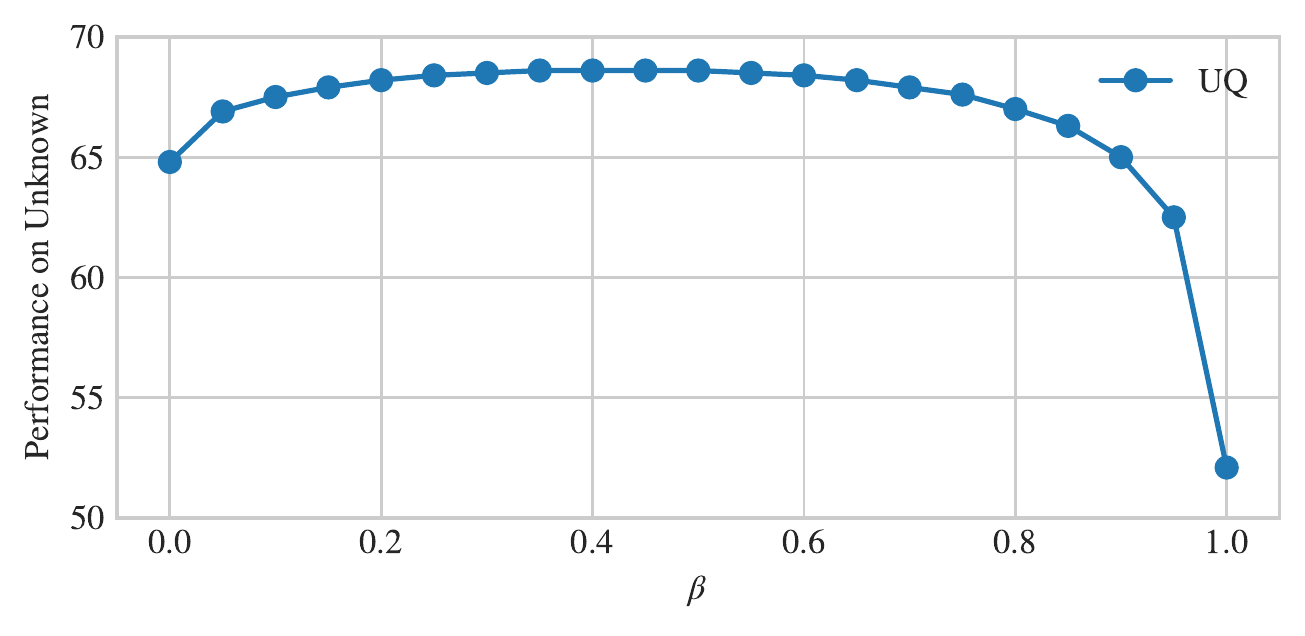}%
}{%
    \caption{
        Ablation study on varying the relative weight $ \beta $
        of using 3D location distance versus embedding distance
        for identifying unknown instances.
    }
    \label{fig:ablation-beta}
}
\end{floatrow}
\vspace{-0.1in}
\end{figure}

\paragraph{Effectiveness of instance-aware embeddings:}
We also study the effectiveness of using instance-aware embeddings
to group unknown points into instances.
In particular, we compare our embeddings against other per-point features,
namely 3D location (\textbf{Points}),
predicted instance center (\textbf{Center}),
and semantic features (\textbf{Semantics}).
From Tab.~\ref{tab:ablation-features}, we see that our instance-aware embeddings
acheive the best results among the alternatives.
Fig.~\ref{fig:ablation-beta} also indicates that using a combination of
instance-aware embeddings and geometry features yields further improvements.

\section{Conclusion}
We have presented a novel and effective open-set instance segmentation method for point clouds.
In particular, we proposed a deep convolutional neural network to encode points into
a category-agnostic embedding space in which they can be clustered into instances.
As a result, our method is able to perceptually group points into instances, irrespective
of whether they belong to a known or unknown class.
We validate our method on two large-scale self-driving datasets and achieve
state-of-the-art performance in the open-set setting.
In the future we plan to explicitly reason about motion as a cue for better instance segmentation of moving objects.

\newpage
{
\small
\bibliography{egbib}

\begin{thebibliography}{50}
\providecommand{\natexlab}[1]{#1}
\providecommand{\url}[1]{\texttt{#1}}
\expandafter\ifx\csname urlstyle\endcsname\relax
  \providecommand{\doi}[1]{doi: #1}\else
  \providecommand{\doi}{doi: \begingroup \urlstyle{rm}\Url}\fi

\bibitem[Saxena et~al.(2008)Saxena, Driemeyer, and Ng]{novelgrasp}
A.~Saxena, J.~Driemeyer, and A.~Y. Ng.
\newblock Robotic grasping of novel objects using vision.
\newblock \emph{IJRR}, 2008.

\bibitem[Ennaceur and Delacour(1988)]{ennaceur1988new}
A.~Ennaceur and J.~Delacour.
\newblock A new one-trial test for neurobiological studies of memory in rats.
  1: Behavioral data.
\newblock \emph{Behavioural brain research}, 31\penalty0 (1):\penalty0 47--59,
  1988.

\bibitem[Antunes and Biala(2012)]{antunes2012novel}
M.~Antunes and G.~Biala.
\newblock The novel object recognition memory: neurobiology, test procedure,
  and its modifications.
\newblock \emph{Cognitive processing}, 2012.

\bibitem[{Scheirer} et~al.(2013){Scheirer}, {de Rezende Rocha}, {Sapkota}, and
  {Boult}]{openworld}
W.~J. {Scheirer}, A.~{de Rezende Rocha}, A.~{Sapkota}, and T.~E. {Boult}.
\newblock Toward open set recognition.
\newblock \emph{TPAMI}, 2013.

\bibitem[Bendale and Boult(2016)]{bendale2016towards}
A.~Bendale and T.~E. Boult.
\newblock Towards open set deep networks.
\newblock In \emph{CVPR}, 2016.

\bibitem[Palmer(2002)]{palmer2002organizing}
S.~E. Palmer.
\newblock Organizing objects and scenes.
\newblock \emph{Foundations of cognitive psychology: Core readings}, 2002.

\bibitem[Palmer(1999)]{palmer1999vision}
S.~E. Palmer.
\newblock \emph{Vision science: Photons to phenomenology}.
\newblock MIT press, 1999.

\bibitem[Shi and Malik(2000)]{ncut}
J.~Shi and J.~Malik.
\newblock Normalized cuts and image segmentation.
\newblock \emph{{IEEE} Trans. Pattern Anal. Mach. Intell.}, 22\penalty0
  (8):\penalty0 888--905, 2000.

\bibitem[Felzenszwalb and Huttenlocher(2004)]{graphbased}
P.~F. Felzenszwalb and D.~P. Huttenlocher.
\newblock Efficient graph-based image segmentation.
\newblock \emph{International Journal of Computer Vision}, 59\penalty0
  (2):\penalty0 167--181, 2004.

\bibitem[Everingham et~al.(2010)Everingham, Gool, Williams, Winn, and
  Zisserman]{voc}
M.~Everingham, L.~J.~V. Gool, C.~K.~I. Williams, J.~M. Winn, and A.~Zisserman.
\newblock The pascal visual object classes {(VOC)} challenge.
\newblock \emph{IJCV}, 88\penalty0 (2):\penalty0 303--338, 2010.

\bibitem[Lin et~al.(2014)Lin, Maire, Belongie, Hays, Perona, Ramanan,
  Doll{\'{a}}r, and Zitnick]{mscoco}
T.~Lin, M.~Maire, S.~J. Belongie, J.~Hays, P.~Perona, D.~Ramanan,
  P.~Doll{\'{a}}r, and C.~L. Zitnick.
\newblock Microsoft {COCO:} common objects in context.
\newblock In \emph{ECCV}, 2014.

\bibitem[Cordts et~al.(2016)Cordts, Omran, Ramos, Rehfeld, Enzweiler, Benenson,
  Franke, Roth, and Schiele]{cityscapes}
M.~Cordts, M.~Omran, S.~Ramos, T.~Rehfeld, M.~Enzweiler, R.~Benenson,
  U.~Franke, S.~Roth, and B.~Schiele.
\newblock The cityscapes dataset for semantic urban scene understanding.
\newblock In \emph{CVPR}, 2016.

\bibitem[Kirillov et~al.(2019)Kirillov, He, Girshick, Rother, and
  Doll{\'{a}}r]{panoptic}
A.~Kirillov, K.~He, R.~B. Girshick, C.~Rother, and P.~Doll{\'{a}}r.
\newblock Panoptic segmentation.
\newblock In \emph{CVPR}, 2019.

\bibitem[Liu et~al.(2019)Liu, Miao, Zhan, Wang, Gong, and Yu]{longtail}
Z.~Liu, Z.~Miao, X.~Zhan, J.~Wang, B.~Gong, and S.~X. Yu.
\newblock Large-scale long-tailed recognition in an open world.
\newblock In \emph{CVPR}, 2019.

\bibitem[Lee et~al.(2018)Lee, Lee, Lee, and Shin]{confcalib}
K.~Lee, H.~Lee, K.~Lee, and J.~Shin.
\newblock Training confidence-calibrated classifiers for detecting
  out-of-distribution samples.
\newblock In \emph{ICLR}, 2018.

\bibitem[Frome et~al.(2013)Frome, Corrado, Shlens, Bengio, Dean, Ranzato, and
  Mikolov]{devise}
A.~Frome, G.~S. Corrado, J.~Shlens, S.~Bengio, J.~Dean, M.~Ranzato, and
  T.~Mikolov.
\newblock Devise: {A} deep visual-semantic embedding model.
\newblock In \emph{NIPS}, 2013.

\bibitem[Socher et~al.(2013)Socher, Ganjoo, Manning, and Ng]{zslcrossmodal}
R.~Socher, M.~Ganjoo, C.~D. Manning, and A.~Y. Ng.
\newblock Zero-shot learning through cross-modal transfer.
\newblock In \emph{NIPS}, 2013.

\bibitem[Xian et~al.(2016)Xian, Akata, Sharma, Nguyen, Hein, and
  Schiele]{latentzsl}
Y.~Xian, Z.~Akata, G.~Sharma, Q.~N. Nguyen, M.~Hein, and B.~Schiele.
\newblock Latent embeddings for zero-shot classification.
\newblock In \emph{CVPR}, 2016.

\bibitem[Pham et~al.(2018)Pham, Kumar, Do, Carneiro, and Reid]{bayesseg}
T.~Pham, B.~G.~V. Kumar, T.~Do, G.~Carneiro, and I.~D. Reid.
\newblock Bayesian semantic instance segmentation in open set world.
\newblock In \emph{ECCV}, 2018.

\bibitem[He et~al.(2017)He, Gkioxari, Doll{\'{a}}r, and Girshick]{maskrcnn}
K.~He, G.~Gkioxari, P.~Doll{\'{a}}r, and R.~B. Girshick.
\newblock Mask {R-CNN}.
\newblock In \emph{ICCV}, 2017.

\bibitem[Arbelaez et~al.(2011)Arbelaez, Maire, Fowlkes, and Malik]{ucm}
P.~Arbelaez, M.~Maire, C.~C. Fowlkes, and J.~Malik.
\newblock Contour detection and hierarchical image segmentation.
\newblock \emph{{IEEE} Trans. Pattern Anal. Mach. Intell.}, 33\penalty0
  (5):\penalty0 898--916, 2011.

\bibitem[Hu et~al.(2018)Hu, Doll{\'{a}}r, He, Darrell, and
  Girshick]{segeverything}
R.~Hu, P.~Doll{\'{a}}r, K.~He, T.~Darrell, and R.~B. Girshick.
\newblock Learning to segment every thing.
\newblock In \emph{CVPR}, 2018.

\bibitem[Krishna et~al.(2017)Krishna, Zhu, Groth, Johnson, Hata, Kravitz, Chen,
  Kalantidis, Li, Shamma, Bernstein, and Fei{-}Fei]{visualgenome}
R.~Krishna, Y.~Zhu, O.~Groth, J.~Johnson, K.~Hata, J.~Kravitz, S.~Chen,
  Y.~Kalantidis, L.~Li, D.~A. Shamma, M.~S. Bernstein, and L.~Fei{-}Fei.
\newblock Visual genome: Connecting language and vision using crowdsourced
  dense image annotations.
\newblock \emph{IJCV}, 123\penalty0 (1):\penalty0 32--73, 2017.

\bibitem[Osep et~al.(2019)Osep, Voigtlaender, Weber, Luiten, and
  Leibe]{generic4d}
A.~Osep, P.~Voigtlaender, M.~Weber, J.~Luiten, and B.~Leibe.
\newblock 4d generic video object proposals.
\newblock \emph{CoRR}, abs/1901.09260, 2019.

\bibitem[Trevor et~al.(2013)Trevor, Gedikli, Rusu, and
  Christensen]{trevor2013efficient}
A.~J. Trevor, S.~Gedikli, R.~B. Rusu, and H.~I. Christensen.
\newblock Efficient organized point cloud segmentation with connected
  components.
\newblock \emph{Semantic Perception Mapping and Exploration (SPME)}, 2013.

\bibitem[Ren and Zemel(2017)]{recattend}
M.~Ren and R.~S. Zemel.
\newblock End-to-end instance segmentation with recurrent attention.
\newblock In \emph{CVPR}, 2017.

\bibitem[Xiong et~al.(2019)Xiong, Liao, Zhao, Hu, Bai, Yumer, and
  Urtasun]{upsnet}
Y.~Xiong, R.~Liao, H.~Zhao, R.~Hu, M.~Bai, E.~Yumer, and R.~Urtasun.
\newblock Upsnet: {A} unified panoptic segmentation network.
\newblock In \emph{CVPR}, 2019.

\bibitem[Chen et~al.(2019)Chen, Girshick, He, and Doll{\'{a}}r]{tensormask}
X.~Chen, R.~B. Girshick, K.~He, and P.~Doll{\'{a}}r.
\newblock Tensormask: {A} foundation for dense object segmentation.
\newblock \emph{CoRR}, abs/1903.12174, 2019.

\bibitem[Pinheiro et~al.(2015)Pinheiro, Collobert, and Doll{\'{a}}r]{deepmask}
P.~H.~O. Pinheiro, R.~Collobert, and P.~Doll{\'{a}}r.
\newblock Learning to segment object candidates.
\newblock In \emph{NIPS}, pages 1990--1998, 2015.

\bibitem[Pinheiro et~al.(2016)Pinheiro, Lin, Collobert, and
  Doll{\'{a}}r]{sharpmask}
P.~O. Pinheiro, T.~Lin, R.~Collobert, and P.~Doll{\'{a}}r.
\newblock Learning to refine object segments.
\newblock In \emph{ECCV}, 2016.

\bibitem[Song and Xiao(2016)]{deepsliding}
S.~Song and J.~Xiao.
\newblock Deep sliding shapes for amodal 3d object detection in {RGB-D} images.
\newblock In \emph{CVPR}, 2016.

\bibitem[Qi et~al.(2018)Qi, Liu, Wu, Su, and Guibas]{fpointnet}
C.~R. Qi, W.~Liu, C.~Wu, H.~Su, and L.~J. Guibas.
\newblock Frustum pointnets for 3d object detection from {RGB-D} data.
\newblock In \emph{CVPR}, 2018.

\bibitem[Bai and Urtasun(2017)]{dwt}
M.~Bai and R.~Urtasun.
\newblock Deep watershed transform for instance segmentation.
\newblock In \emph{CVPR}, 2017.

\bibitem[Liang et~al.(2018)Liang, Lin, Wei, Shen, Yang, and Yan]{proposalfree}
X.~Liang, L.~Lin, Y.~Wei, X.~Shen, J.~Yang, and S.~Yan.
\newblock Proposal-free network for instance-level object segmentation.
\newblock \emph{{IEEE} Trans. Pattern Anal. Mach. Intell.}, 40\penalty0
  (12):\penalty0 2978--2991, 2018.

\bibitem[Neven et~al.(2019)Neven, Brabandere, Proesmans, and
  Gool]{jointbandwidth}
D.~Neven, B.~D. Brabandere, M.~Proesmans, and L.~V. Gool.
\newblock Instance segmentation by jointly optimizing spatial embeddings and
  clustering bandwidth.
\newblock In \emph{CVPR}, 2019.

\bibitem[Liu et~al.(2017)Liu, Jia, Fidler, and Urtasun]{sgn}
S.~Liu, J.~Jia, S.~Fidler, and R.~Urtasun.
\newblock {SGN:} sequential grouping networks for instance segmentation.
\newblock In \emph{ICCV}, 2017.

\bibitem[Wang et~al.(2018)Wang, Yu, Huang, and Neumann]{sgpn}
W.~Wang, R.~Yu, Q.~Huang, and U.~Neumann.
\newblock {SGPN:} similarity group proposal network for 3d point cloud instance
  segmentationeccv.
\newblock In \emph{CVPR}, 2018.

\bibitem[Brabandere et~al.(2017)Brabandere, Neven, and Gool]{pushpull}
B.~D. Brabandere, D.~Neven, and L.~V. Gool.
\newblock Semantic instance segmentation with a discriminative loss function.
\newblock \emph{CoRR}, abs/1708.02551, 2017.

\bibitem[Liu and Furukawa(2019)]{masc}
C.~Liu and Y.~Furukawa.
\newblock {MASC:} multi-scale affinity with sparse convolution for 3d instance
  segmentation.
\newblock \emph{CoRR}, abs/1902.04478, 2019.

\bibitem[Pham et~al.(2019)Pham, Nguyen, Hua, Roig, and Yeung]{jsis}
Q.~Pham, D.~T. Nguyen, B.~Hua, G.~Roig, and S.~Yeung.
\newblock {JSIS3D:} joint semantic-instance segmentation of 3d point clouds
  with multi-task pointwise networks and multi-value conditional random fields.
\newblock In \emph{CVPR}, 2019.

\bibitem[Elich et~al.(2019)Elich, Engelmann, Schult, Kontogianni, and
  Leibe]{bevis}
C.~Elich, F.~Engelmann, J.~Schult, T.~Kontogianni, and B.~Leibe.
\newblock {3D-BEVIS:} birds-eye-view instance segmentation.
\newblock \emph{CoRR}, abs/1904.02199, 2019.

\bibitem[Fathi et~al.(2017)Fathi, Wojna, Rathod, Wang, Song, Guadarrama, and
  Murphy]{metricseg}
A.~Fathi, Z.~Wojna, V.~Rathod, P.~Wang, H.~O. Song, S.~Guadarrama, and K.~P.
  Murphy.
\newblock Semantic instance segmentation via deep metric learning.
\newblock \emph{CoRR}, abs/1703.10277, 2017.

\bibitem[Kong and Fowlkes(2018)]{recurrentemb}
S.~Kong and C.~C. Fowlkes.
\newblock Recurrent pixel embedding for instance grouping.
\newblock In \emph{CVPR}, pages 9018--9028, 2018.

\bibitem[Cheng(1995)]{meanshift}
Y.~Cheng.
\newblock Mean shift, mode seeking, and clustering.
\newblock \emph{{IEEE} Trans. Pattern Anal. Mach. Intell.}, 17\penalty0
  (8):\penalty0 790--799, 1995.

\bibitem[Ester et~al.(1996)Ester, Kriegel, Sander, and Xu]{dbscan}
M.~Ester, H.-P. Kriegel, J.~Sander, and X.~Xu.
\newblock Density-based spatial clustering of applications with noise.
\newblock In \emph{KDD}, 1996.

\bibitem[Tchapmi et~al.(2017)Tchapmi, Choy, Armeni, Gwak, and
  Savarese]{segcloud}
L.~Tchapmi, C.~Choy, I.~Armeni, J.~Gwak, and S.~Savarese.
\newblock Segcloud: Semantic segmentation of 3d point clouds.
\newblock In \emph{3DV}, 2017.

\bibitem[Zhang et~al.(2018)Zhang, Luo, and Urtasun]{chris}
C.~Zhang, W.~Luo, and R.~Urtasun.
\newblock Efficient convolutions for real-time semantic segmentation of 3d
  point clouds.
\newblock In \emph{3DV}, 2018.

\bibitem[Yang et~al.(2018)Yang, Liang, and Urtasun]{hdnet}
B.~Yang, M.~Liang, and R.~Urtasun.
\newblock Hdnet: Exploiting hd maps for 3d object detection.
\newblock In \emph{CoRL}, 2018.

\bibitem[Snell et~al.(2017)Snell, Swersky, and Zemel]{protonet}
J.~Snell, K.~Swersky, and R.~S. Zemel.
\newblock Prototypical networks for few-shot learning.
\newblock In \emph{NIPS}, 2017.

\bibitem[Kingma and Ba(2015)]{adam}
D.~P. Kingma and J.~Ba.
\newblock Adam: {A} method for stochastic optimization.
\newblock In \emph{ICLR}, 2015.

\end{thebibliography}
}

\newpage
{
\appendix

\section{Additional Results}
\begin{figure}[!htb]
    \includegraphics[width=\textwidth]{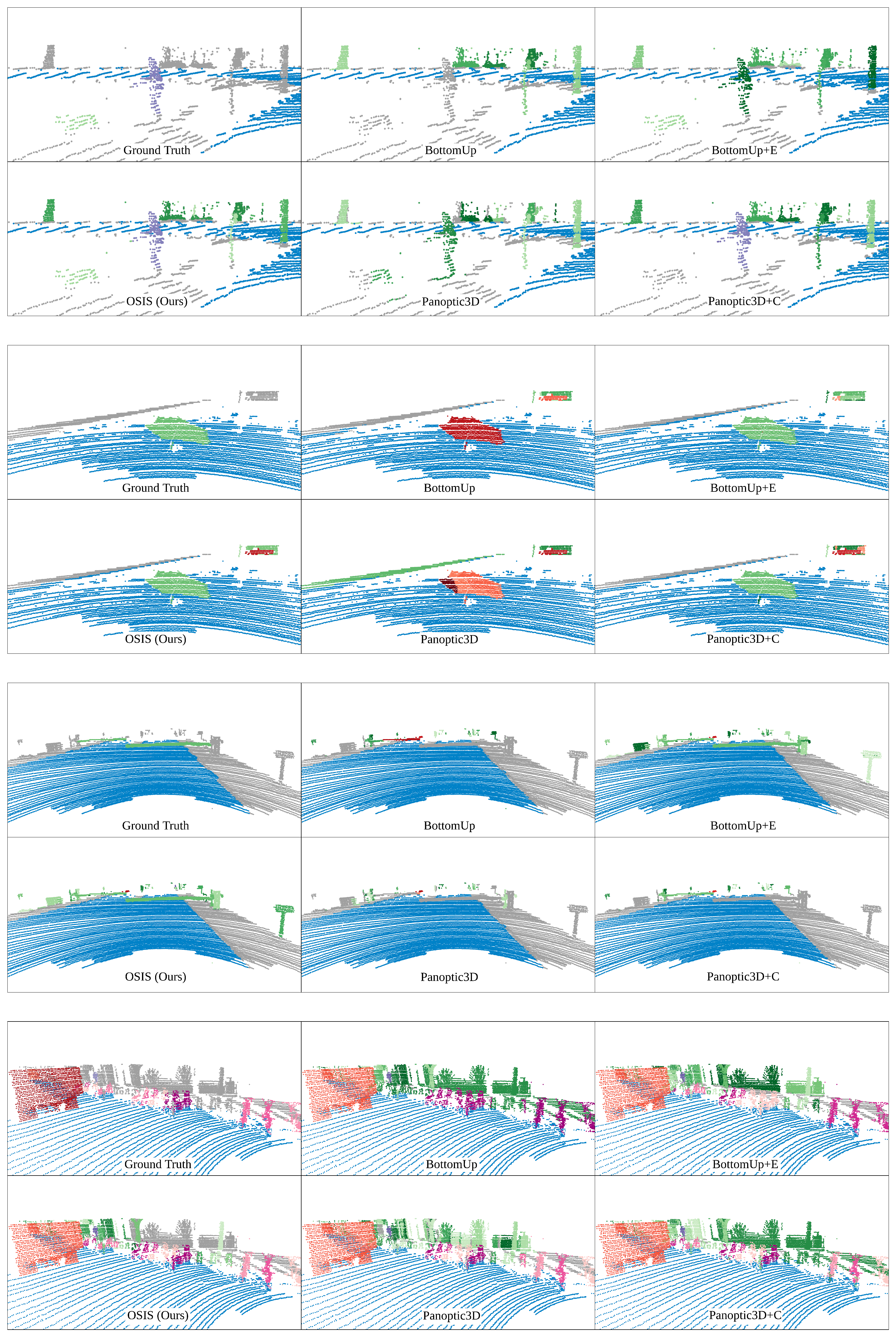}
    \caption{Qualitative results of open-set instance segmentation on TOR4D and Rare4D.}
\end{figure}

}

\end{document}